\documentclass{article}

\usepackage{arxiv}

\usepackage[utf8]{inputenc} 
\usepackage[T1]{fontenc}    
\usepackage{hyperref}       
\usepackage{url}            
\usepackage{booktabs}       
\usepackage{amsfonts}       
\usepackage{nicefrac}       
\usepackage{microtype}      
\usepackage{cleveref}       
\usepackage{lipsum}         
\usepackage{graphicx}
\usepackage{natbib}
\usepackage{doi}
\usepackage{comment}

\title{Hardware Trojan Insertion Using Reinforcement Learning}

\date{}

\author{ \href{https://orcid.org/my-orcid?orcid=0000-0002-0134-8418}{\includegraphics[scale=0.06]{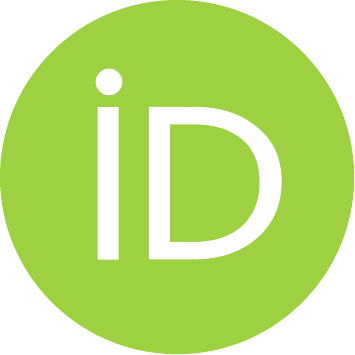}\hspace{1mm}Amin ~Sarihi}, \\
	Klipsch School of Electrical\\ and Computer Engineering\\
	New Mexico State University\\
	\texttt{sarihi@nmsu.edu} \\
	\And
	\href{https://orcid.org/0000-0003-2647-2797}{\includegraphics[scale=0.06]{orcid.pdf}\hspace{1mm}Ahmad ~Patooghy} \\
	Computer Systems Technology\\
	North Carolina A\&T  University\\
	\texttt{apatooghy@ncat.edu}\\
	\And
	\href{https://orcid.org/0000-0002-3741-0201}{\includegraphics[scale=0.06]{orcid.pdf}\hspace{1mm}Peter ~Jamieson} \\
	Department of Electrical\\ and Computer Engineering\\
	Miami University\\
	\texttt{jamiespa@miamioh.edu}\\
	\And
	\href{https://orcid.org/0000-0001-8027-1449}{\includegraphics[scale=0.06]{orcid.pdf}\hspace{1mm}Abdel-Hameed A. ~Badawy} \\
	Klipsch School of Electrical\\ and Computer Engineering\\
	New Mexico State University\\
	\texttt{badawy@nmsu.edu} \\}

\renewcommand{\shorttitle}{Hardware Trojan Insertion Using Reinforcement Learning}

\begin{document}
\maketitle

\begin{abstract}
	This paper utilizes Reinforcement Learning (RL) as a means to automate the Hardware Trojan (HT) insertion process to eliminate the inherent human biases that limit the development of robust HT detection methods. An RL agent explores the design space and finds circuit locations that are best for keeping inserted HTs hidden. To achieve this, a digital circuit is converted to an environment in which an RL agent inserts HTs such that the cumulative reward is maximized. Our toolset can insert combinational HTs into the ISCAS-85 benchmark suite with variations in HT size and triggering conditions. Experimental results show that the toolset achieves high input coverage rates (100\% in two benchmark circuits) that confirms its effectiveness. Also, the inserted HTs have shown a minimal footprint and rare activation probability.
\end{abstract}


\section{Introduction}
 
With recent advancements in VLSI technologies, integrated circuits (ICs) manufacturing involves multiple companies, introducing new security challenges at each stage of IC production. 
Hardware Trojans (HTs) are defined as any undesired modification in an IC which can lead to erroneous outputs and or leak of information~\cite{pan2021automated}. According to the adversarial model introduced by~\cite{shakya2017benchmarking}, HTs can be inserted into the target IC in three different scenarios, namely intellectual properties (IPs) (processing cores, various I/O components, and network-on-chip~\cite{sarihi2021survey}), disgruntled employees at the integration stage, reverse-engineering by an untrusted foundry.

To study the behavior of HTs in digital circuits, researchers have been mostly using limited benchmarks ~\cite{shakya2017benchmarking,salmani2013design} including a set of 91 trust benchmarks with different HT sizes and configurations (available at~\url{trust-hub.org}). Over the past years, various HT detection approaches have been developed based on these benchmarks~\cite{salmani2016cotd, sabri2021sat,hasegawa2017trojan, sebt2018circuit}. Despite the valuable effort to create such HT benchmarks, they are limited in size and variety necessary to push the detection tools. Another downside of using these existing benchmarks is the problem of dealing with fixed static trigger conditions~\cite{cruz2018automated}, which stems from the human bias during HT insertions. As a result, HT detectors can be tuned in a way to enhance their detection accuracy while not truly being effective in complex and real-world modern ICs. These shortcomings emphasize the need for an automated HT-insertion tool free of human biases that can be used to create high-volume HT benchmarks. Such a tool will help implement HT benchmarks aligned with the fast growth of attack approaches and cater to the security needs in the realm of hardware design. By introducing new HTs into a design, one can create new benchmarks and push the capabilities of HT detection via introducing never-seen-before HTs. 

Although a few researchers have tried to address these problems by introducing tunable HT insertion toolsets~\cite{cruz2018automated,yu2019improved}, these approaches have no concrete guideline for selecting trigger and payload nets; instead, triggering is done on an ad-hoc basis with little design space exploration capabilities. In this paper, we attempt to develop an HT-insertion tool, free of human biases, using a Reinforcement Learning (RL) agent that decides where to insert HTs through a trial and error method. Although machine learning techniques have been used to detect HTs in the past~\cite{xue2020ten,salmani2016cotd,hasegawa2017trojan}, to the best of our knowledge, this work is the first that addresses HT insertion using a machine learning approach via design space exploration.

Our toolset translates each circuit to a graph representation in which different properties of each net, such as controllability, observability, and logical depth, are computed (so-called the SCOAP\footnote{Sandia controllability and observability analysis program.} parameters ~\cite{goldstein1980scoap}). The circuit graph is considered as an environment in which the RL agent tries to insert the HT to maximize the gained rewards. 
Obtained results confirm that the inserted HTs are very hard to detect as the toolset maximizes the number of IC's inputs involved in the activation of the inserted HTs. We define a metric called the input coverage percentage (ICP) to determine the difficulty of HT activation.

The paper is organized as follows: The mechanics of our proposed approach are presented in Section~\ref{proposed}. Section~\ref{results} demonstrates the experimental results and Section~\ref{conclusion} concludes the paper.
\section{RL-based HT insertion}
\label{proposed}
\begin{figure*}[ht]
  \centering
  \includegraphics[scale=.48]{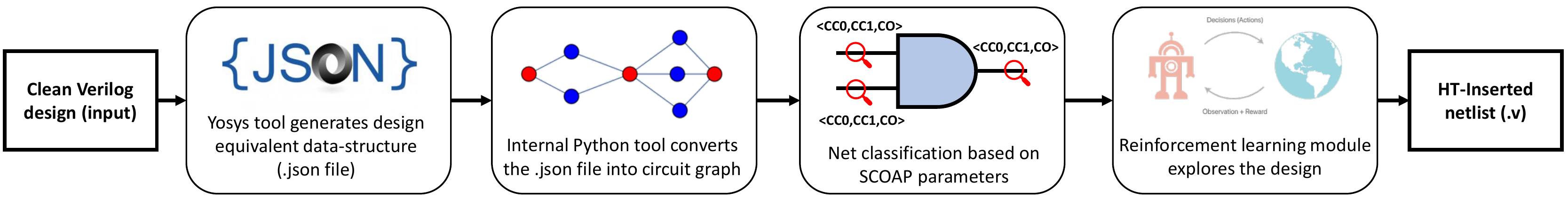}
  \caption{The proposed RL-based HT insertion tool flow.}
  \label{Toolflow}
\end{figure*}
Figure~\ref{Toolflow} shows the flow of the proposed HT insertion tool. The first step to insert an HT into a circuit is to create a graph representation of the flattened netlist from the circuit. 
Yosys Open Synthesis Suite~\cite{wolf2013yosys} translates a Verilog file of the circuit into a JSON (JavaScript Object Notation)~\cite{bassett2015introduction} netlist where the JSON file is used by a python script to parse the internal graph representation of the circuit. Next, the tool finds a set of rare nets to be used as HT trigger nets (this is described in detail in Subsection~\ref{s1}). Finally, an RL agent uses the rare net information and attempts to insert an HT to maximize a reward function as described in Subsection~\ref{s2}.

\subsection{Rare Nets Extraction}
\label{s1}

As discussed earlier, different circuit criteria have been used for trigger selection. In this work, we use the parameters introduced in \cite{sebt2018circuit} where the trigger nets are selected based on functions of net \emph{controllability} and \emph{observability} \cite{goldstein1980scoap}. 

The first parameter is called the HT trigger susceptibility parameter, and it is derived from the fact that low switching nets have a high difference between their controllability values. Equation \ref{HTS1} describes this parameter:
\begin{equation}
    HTS(Net_i)=\frac{|CC1(Net_i)-CC0(Net_i)|}{Max(CC1(Net_i),CC0(Net_i))}
    \label{HTS1}
\end{equation}
where $HTS$ is the HT trigger susceptibility parameter of a net; $CC0(Net_i)$ and $CC1(Net_i)$ are the combinational controllability 0 and 1 of $Net_i$, respectively. The $HTS$ parameter ranges between $[0,1)$ such that higher values of $HTS$ are correlated with lower activity on the net. 

The other used parameter is specified in Equation~\ref{OCR} to measure the ratio of observability to controllability:
\begin{equation}
    OCR(Net_i)=\frac{CO(Net_i)}{CC1(Net_i)+CC0(Net_i)}
    \label{OCR}
\end{equation}
where $OCR$ is the observability to controllability ratio of a net. This equation requires that the HT trigger nets must be very hard to control but not so hard to observe. Unlike the $HTS$ parameter, $OCR$ is not bounded, and it belongs to the interval of $[0,\infty)$. We will specify thresholds (see Section~\ref{s2}) for each parameter and use them as filters to populate the set of rarely-activated nets for our tool.
\subsection{RL-Based HT Insertion.}
\label{s2}

Agent, Action, Environment, State, and Reward are the five main components of reinforcement learning. From an RL perspective, we define the environment as the circuit in which we are trying to insert HTs. The agent's action is the insertion of the HT. Note, we consider combinational HTs where trigger nets are ANDED, and the payload is an XOR gate. 

We represent different HT insertions with a state vector in each circuit. To address this issue, we first levelize the circuit. The output level of an $m$-input gate is computed by equation~\ref{eq1}:
\begin{equation}
    Level(out)=MAX(Level(in_1), Level(in_2), ... , Level(in_m))+1
    \label{eq1}
\end{equation}

\begin{figure}[!t]
  \centering
  \includegraphics[width=3.3in, height=1.6in]{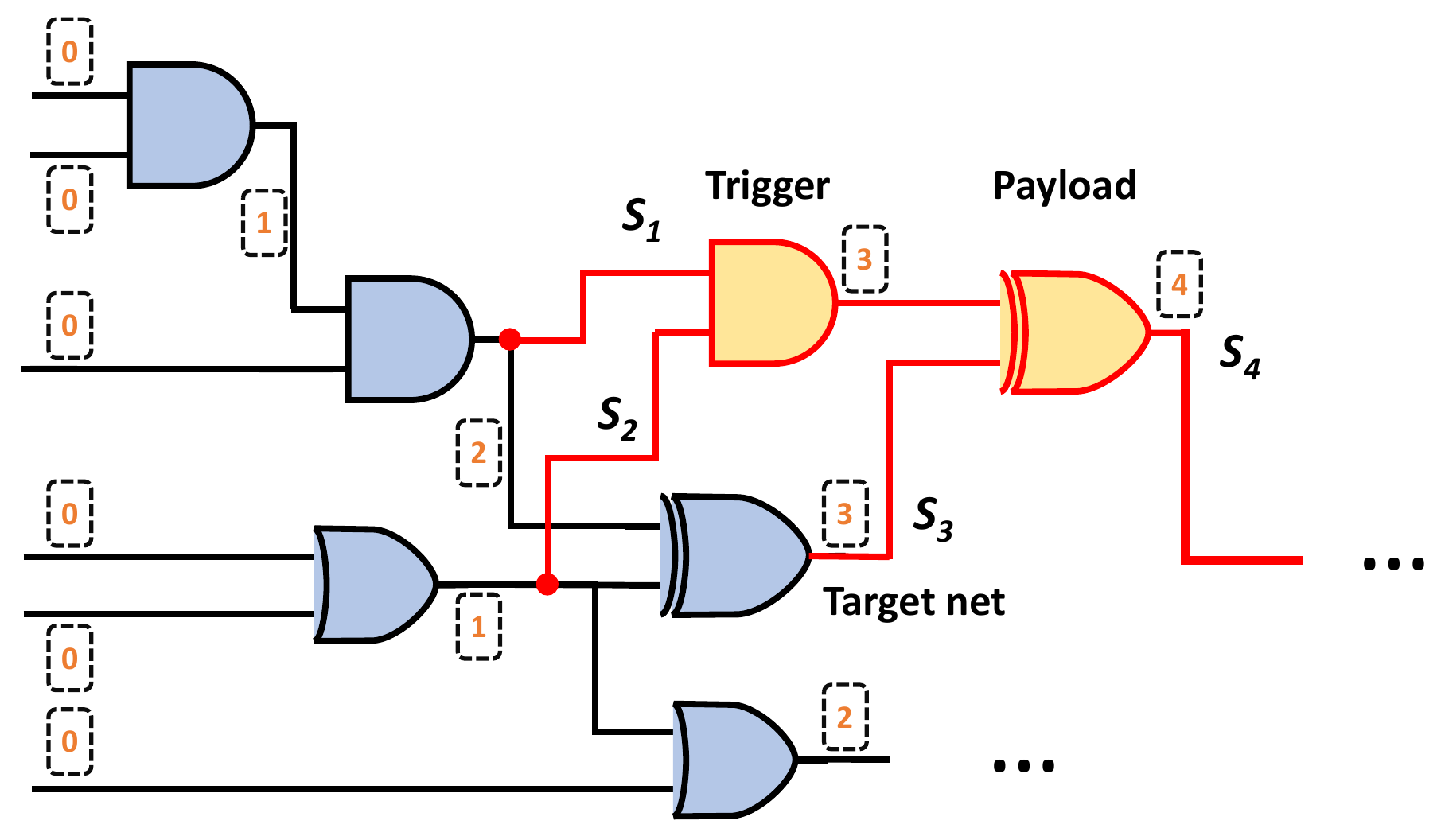}
  \caption{Obtaining the state vector in the presence of an HT.}
  \label{State}
\end{figure}

Figure~\ref{State} depicts a 2-input HT (in yellow) where the XOR payload flips the value of the target net when the trigger is activated. For a given HT, the state vector is comprised of $s_t=[s_1,s_2, ...,s_{n-2},s_{n-1},s_{n}]$ where $s_1$ through $s_{n-2}$ are levels of the HT inputs and $s_{n-1}$ and $s_{n}$ are levels of the target net and the output of the XOR payload, respectively. As an example, the HT in Figure~\ref{State} has the state vector $s_t=[2,1,3,4]$. The action space of the described HT agent is multi-discrete, i.e, each input of the HT can choose an action from a set of 5 available actions. These actions are:

\begin{itemize}
    \item \emph{\textbf{Next level}}: the input of the HT moves to one of the nets that are one level higher than the current net level.
    \item \emph{\textbf{Previous level}}: the input of the HT moves to one of the nets that are one level lower than the current net level.
    \item \emph{\textbf{Same level up}}: the input of the HT will move to one of the nets at the same level as the current net level. The net is picked by pointing to the next net in the ascending list of net ids for the given level. 
    \item \emph{\textbf{Same level down}}: the input of the HT will move to one of the nets at the same level as the current net level. The net is picked by pointing to the previous net in the ascending list of nets for the given level. 
    \item \emph{\textbf{No action}}: the input of the HT will not move.
\end{itemize}

If an action leads the agent to step outside the circuit boundaries, it is substituted with a ``No action''.

The action space is also represented by a vector where its size is equal to the number of the HT inputs, and each action can be one of the five actions above, e.g., for the HT in Figure~\ref{State}, the action space would be $a_t=[a_1,a_2]$ since it has two inputs. 

As we explained in Section~\ref{s1}, SCOAP parameters are first computed. We specify two thresholds $T_{HTS}$ and $T_{OCR}$ and require our algorithm to find nets that have higher $HTS$ values than $T_{HTS}$ and lower $OCR$ values than $T_{OCR}$. These nets are classified as suspicious nets. 

Our toolset utilizes an algorithm that consists of two conditional while loops that keep track of the terminal states and the elapsed timesteps. The first used function in the algorithm is called \emph{reset\_environment} which resets the environment before each episode. Upon reset, an HT is randomly inserted within the circuit according to the following set of rules.
\begin{itemize}
    \item Rule 1) Trigger nets are selected randomly from the list of the total nets.
    \item Rule 2) No trigger net is allowed to be fed from a previously used net.
    \item Rule 3) Trigger nets cannot be assigned as the target.
    \item Rule 4) The target net is selected considering the level of trigger nets. To prevent forming combinational loops, we specify that the level of the target net should be greater than that of the trigger nets. 
\end{itemize}

During the training process in each episode, we do not change the payload net to help the RL algorithm converge faster for possible solutions
. Unlike the manual payload selection, we allow the algorithm explore the environment during different episodes and decide which payload can more seriously compromise the security by collecting higher rewards. This solution addresses the problem of finding optimal payload selection.


The training process of the agent takes place in a loop where actions are being issued, rewards are collected, the state is updated, and eventually, the updated graph is returned. To evaluate the taken actions by the RL agent (meaning if the HT can be triggered with any input vector), we use PODEM (Path-Oriented Decision Making), an automatic test pattern generator~\cite{bushnell2000essentials}. If the HT payload propagates through at least one of the circuit outputs, the action gains a reward proportional to the number of inputs of the circuit that are engaged in the activation of the HT (we call this feature input coverage). We believe that the number of inputs engaged in the HT activation could be viewed as a metric of how rarely the HT is activated (see Section \ref{results}).

In PODEM, $input\_stack$ is the list of circuit inputs and their values that activate the HT. The more inputs engaged, the higher the reward will be. For an action to get positively rewarded, at least one of the HT trigger nets must belong to the set of the suspicious nets. This type of rewarding encourages the agent to search in the vicinity of the suspicious nets (states) where rewards are more positive and hence result in stealthier HTs. The agent is rewarded -1 when the trigger nets do not belong to the set of suspicious nets. Our proposed RL rewarding scheme drives the agent towards inserting hard-to-active HTs and maximizing the input coverage.
The rewarding scheme is given in Equation \ref{reward}
\begin{equation}
    reward+= 20*(size(input\_stack)/size(in\_ports))
    \label{reward}
\end{equation}
where $in\_ports$ is the total number of inputs.
We selected the coefficient 20 since it strikes a balance between the mostly '-1' rewards collected during training and the limited number of HTs found in each episode. One key benefit of using our tool is that the designer can adjust the reward scheme to achieve different goals. 


To train the RL agent, we use the PPO (Proximal Policy Optimization) RL algorithm. PPO can train agents with muti-discrete action spaces in discrete or continuous spaces. The main idea of PPO is that the updated new policy (which is a set of actions to reach the goal) should not deviate too far from the old policy following an update in the algorithm. To avoid substantial updates, the algorithm uses a technique called clipping in the objective function~\cite{schulman2017proximal}. At last, when the HTs are inserted, the toolset outputs Verilog gate-level netlist files that contain  the malicious HTs.

\section{Experimental Results}
\label{results}
In this section, we demonstrate the efficiency of the developed HT insertion toolset. We look at the number of successful insertions and the achieved input coverage percentage (ICP). The ICP is the ratio of circuit inputs that are engaged in the activation of a successfully inserted HT.  From the attacker's point of view, a higher ICP is desired as it requires more effort for the security engineer to activate the HT and discover it. For instance, an inserted HT with 100\% ICP means that there is only one unique set of inputs that activates the HT.
For our experiments we use an Intel Core i7-5960X CPU @ 3.00GHz and 64 GB of RAM to train and test our agent. The training of the RL agent is done using Stable Baselines library~\cite{raffin2019stable} with MLP (multi-layer perceptron) as the PPO algorithm policy. For testing, the agent is managed to insert HTs into ISCAS-85 benchmarks, which are converted into equivalent circuit graph using NetworkX~\cite{SciPyProceedings_11}.

\begin{table*}[t]

\caption{Number of inserted HTs with different size and percentage of input engagement }
\label{HTS_table}
\begin{center}
 \includegraphics[scale=0.88]{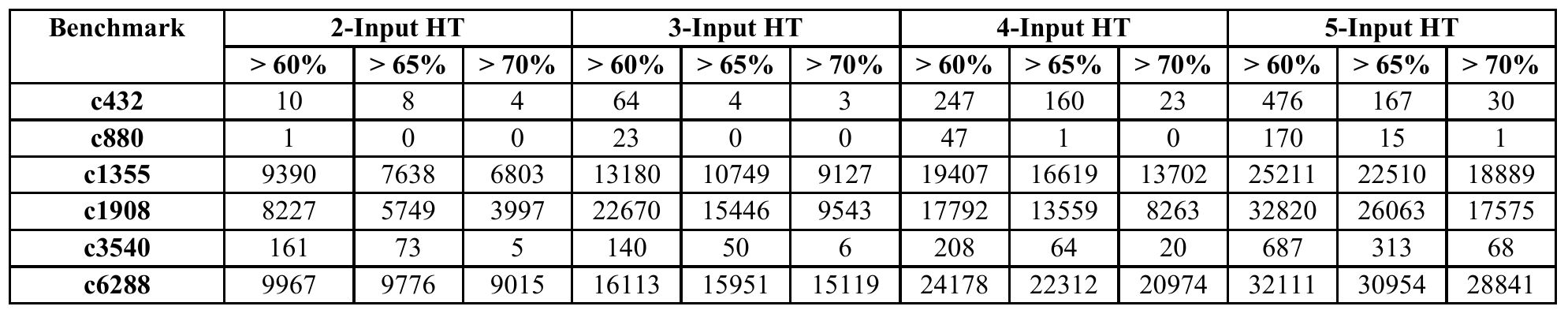}

\end{center}
\end{table*}

\begin{figure}[!t]
  \centering
  \includegraphics[scale=0.42]{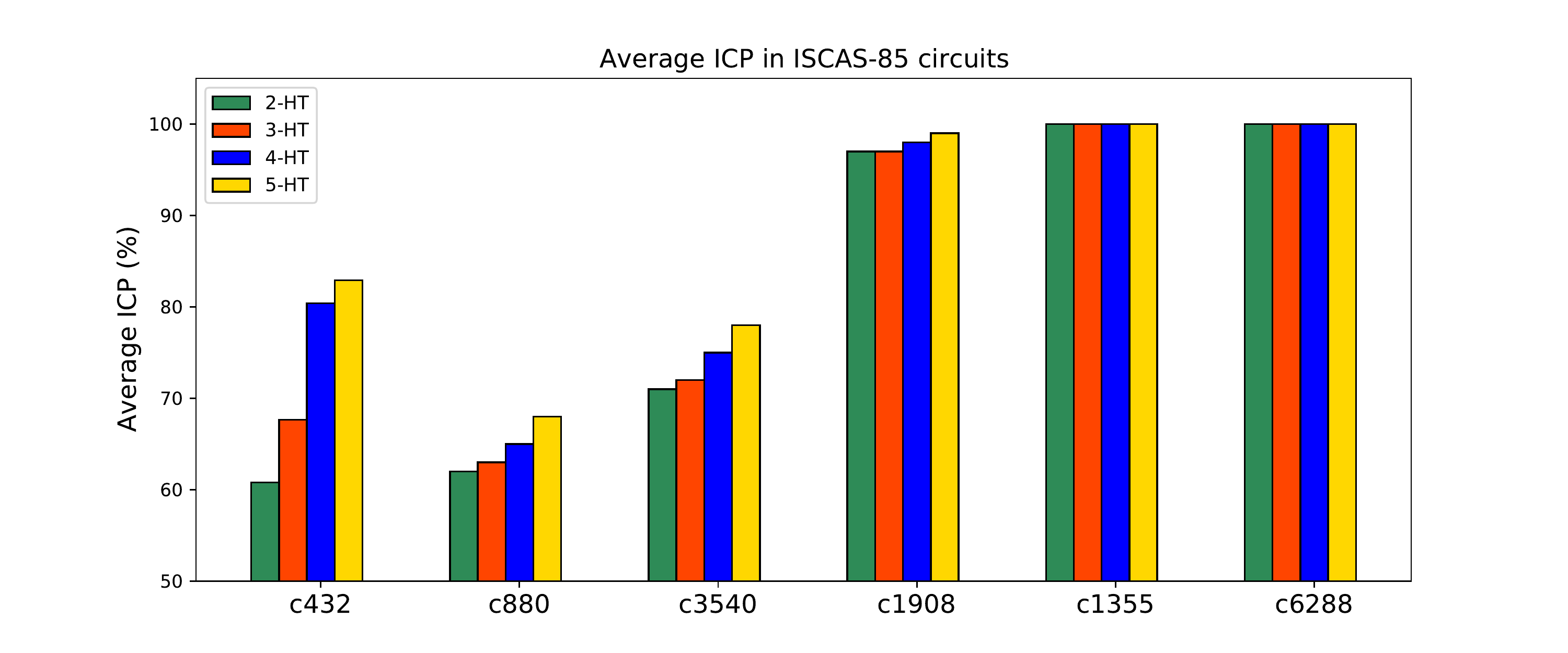}
  \caption{Average ICP of the top 10 inserted HTs.}
  \label{bar}
\end{figure}

\begin{figure}[!t]
  \centering
  \includegraphics[scale=0.45]{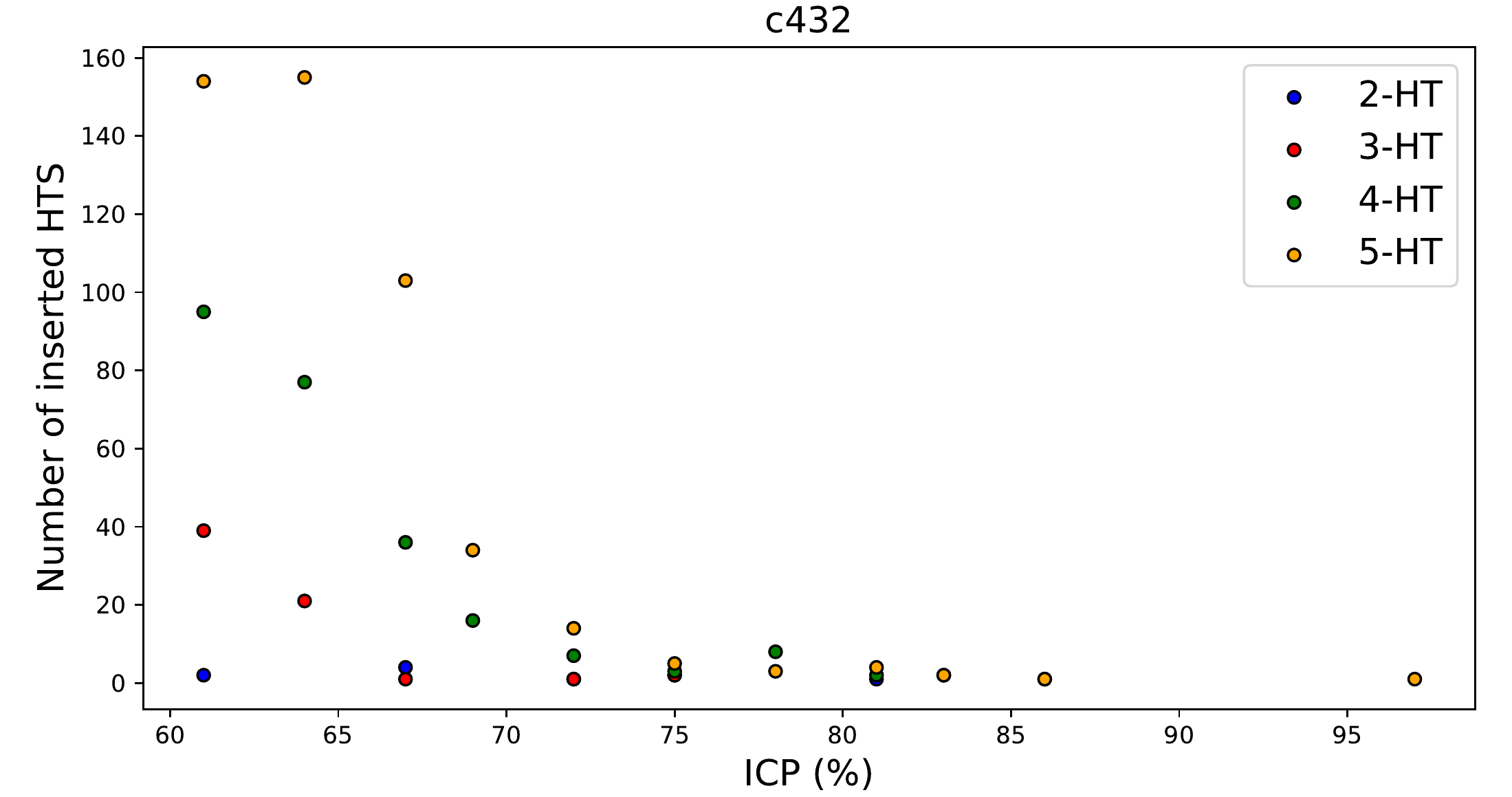}
  \caption{Scatter plot of inserted HTs vs. ICP percentage.}
  \label{scatter}
\end{figure}
For our experiments, we have specified $T_{OCR}$ and $T_{HTS}$ such that 5\% of all nets in each circuit are considered as \emph{suspicious nets}. This was done to enable a fair comparison between the circuits. Also, we increase the number of total timesteps and steps per episode by 10\% every time we increase the HT input size by one. This gives the algorithm more time to explore the environment and keeps the experimental setup relatively consistent as the HT size increases. Our initial total timestep for each circuit is 120000 steps for an HT of size two.

Table~\ref{HTS_table} shows the number of inserted HTs under various ICPs greater than thresholds (60\%, 65\% and 70\%) in six circuits of the ISCAS-85 benchmark for different HT sizes. 
As can be observed in most cases, as the number of HT inputs increases, our tool can insert more HTs with higher ICP into the circuit. This stems from the fact that more HT inputs broadens the possible logic cones that can be activated. Subsequently, more inputs are engaged in the HT activation process. Figure~\ref{bar} verifies the previous statement by comparing the average ICP of the top 10 inserted HTs (in terms of the highest ICP) for all the circuits. As can be seen, as the HT input size increases, the toolset is able to insert HTs with higher ICP in c432, c880, c1908, and c3540. The toolset is able to insert HTs in c1355 and c6288 with an ICP of 100\% even with 2-input HT. This number is just below 100\% in c1908.

Figure~\ref{scatter} depicts the number of inserted HTs against the ICP in c432. The agent was able to insert a 5-input HT that covers 97\% of the inputs while the smaller HTs were unable to achieve such ICP; however, this high ICP comes with the cost of bigger HT in the circuit that could disclose its presence. Interestingly, the 2-input HT has achieved 81\% ICP. Hence the attacker can strike a balance between the HT footprint and its trigger coverage.

The impact of inherent randomness in choosing an Agent action or resetting the environment in the RL algorithm is observed in the results for c3540, where the number of found inserted 3-inputs HTs is less than the 2-input HTs. The results partly rely on what state the agent starts from and what sequence of actions the agent takes. 

Another observation is how different benchmark circuits differ in the overall magnitude of inserted  HTs.  Our hypothesis is that the functionality of a circuit impacts how easy it is to insert HTs. As a case study, we investigate the different characteristics of circuits c880 and c1355. We first compute the average input access of all the target nets in the inserted HTs set, which indicates a path between all the inputs and the target nets. This number is 53\% in c880 while it is 88\% in c1355. Hence, the target nets can access more inputs on average in c1355. Moreover, an analysis of the suspicious nets in both circuits indicates that, unlike c880, there are different sets of nets in c1355 that share the same $HTS$ and $OCR$ parameters, which we call $buses$. There are two 32-bit buses in the suspicious nets set. We observed that the agent only needs one trigger net from these two buses to provide high ICP. Hence, the number of possible inserted HTs requiring only one such net is higher compared to c880. This difference originates from the two circuits' different functionalities and graph structures.

\section{Conclusion}
\label{conclusion}
This paper presented an automated RL HT insertion toolset to address the problem of human bias in HT benchmark design. The rewarding scheme in our study was defined such that the agent finds the HTs that can engage as many inputs as possible for HT activation. The goal was achieved by training an RL agent through trial and error. Unlike the previous work, where normal trigger nets were chosen randomly, our agent automatically picked a set of normal nets alongside the rare nets to maximize the cumulative rewards of the learning agent. This new method of inserting HTs in the ISCAS-85 benchmark circuits can help researchers in the community to devise better HT detection techniques by increasing the HT benchmark set. 

\bibliographystyle{ACM-Reference-Format}
\bibliography{references}

\end{document}